\documentclass[letterpaper, 10 pt, conference]{ieeeconf}  

\IEEEoverridecommandlockouts                              

\usepackage{amsfonts}
\usepackage{booktabs}
\usepackage{soul}
\usepackage{graphicx}
\usepackage{float}
\usepackage{cite}
\usepackage{amsmath}
\usepackage{algorithm,algorithmic}
\usepackage{caption}
\usepackage[dvipsnames]{xcolor}
\usepackage{wrapfig}
\usepackage{hyperref}
\usepackage{subcaption}
\usepackage{flushend}
\usepackage{multicol,multirow}
\newcommand{\squeezeeq}[1]{%
  \begingroup
  \thickmuskip=2mu plus 1mu
  \medmuskip=1mu plus 1mu
  #1
  \endgroup
}

\definecolor{YellowGreen}{rgb}{0.6,0.8,0.2}
\definecolor{LightGray}{gray}{0.92}
\newcommand{\cmark}{\textcolor{ForestGreen}{$\checkmark$}}
\newcommand{\xmark}{\textcolor{Red}{$\times$}}
\hypersetup{
  colorlinks=true,
  citecolor=blue,
  linkcolor=black,
  urlcolor=blue
}
\usepackage{microtype} 

\title{\LARGE \bf
Co-VLA: Consensus-based Federated Training for Vision-Language-Action Models
}

\author{Anonymous Authors}
\author{Haolong Li$^{1}$, Guner Dilsad Er$^{2}$, Michael Muehlebach$^{2}$, Joerg Stueckler$^{1}$
\thanks{*This work has been supported by Hightech Agenda Bayern and by University of Augsburg via the research program “Forschungspotentiale besser nutzen!”. We gratefully acknowledge the HPC resources provided by the Erlangen National High Performance Computing Center (NHR@FAU) of the Friedrich-Alexander-Universität Erlangen-Nürnberg (FAU) under the BayernKI project v119ee. BayernKI funding is provided by Bavarian state authorities.}
\thanks{$^{1}$Haolong Li and Joerg Stueckler are with
the Intelligent Perception in Technical Systems, University of Augsburg, Germany. {\ttfamily\small firstname.lastname@uni-a.de}.}%
\thanks{$^{2}$Guner Dilsad Er and Michael Muehlebach are with
the Max Planck Institute for Intelligent Systems, Germany.}%
}

\begin{document}
\hyphenation{partici-pation}
\bstctlcite{IEEEexample:BSTcontrol} 

\maketitle
\thispagestyle{empty}
\pagestyle{empty}

\begin{abstract}
Vision-language-action models (VLAs) have emerged as a promising paradigm for general-purpose robot learning, with performance improving as models and datasets scale. Scaling robot data collection, however, remains challenging because data are naturally distributed across robots, tasks, and locations, making centralization costly or impractical.
Federated learning offers a way to train on decentralized robot data, but applying it to VLAs requires accounting for heterogeneous robot client data distributions.
We present Co-VLA, which applies consensus optimization using the Alternating Direction Method of Multipliers~(ADMM) to federated VLA training. We show that the same algorithm supports both full-model training and parameter-efficient fine-tuning  with both fixed-rank and rank-adaptive adapters. The name Co-VLA reflects both consensus and collaboration: clients with different local robot datasets collaboratively train a shared model without sharing their data. Our experiments demonstrate that Co-VLA achieves performance comparable to centralized training in both full-model training and parameter-efficient fine-tuning settings. The project website and videos of our real-world experiments are available at
\url{https://embodiedvision.github.io/co-vla/}.
\end{abstract}

\section{Introduction}
\begin{figure*}[t]
    \centering
    \includegraphics[width=0.8\textwidth]{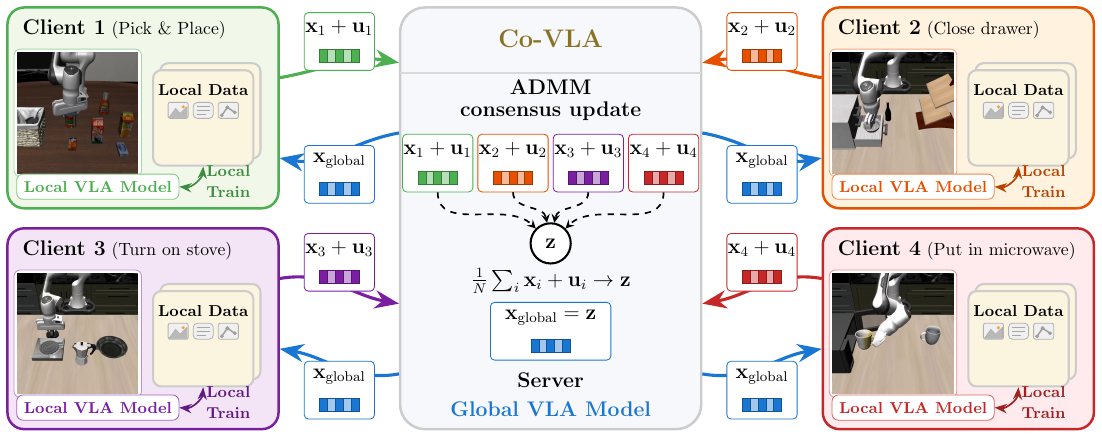}
    \caption{Overview of Co-VLA for federated training of vision-language-action models via global consensus.}
    \label{fig:admm_vla_teaser}
\end{figure*}

Vision-language-action models (VLAs) have emerged as a promising foundation for general-purpose robot learning by mapping visual observations and language instructions directly to robot actions. Recent works~\cite{octo_2023,google2023rt1, zitkovich23a_rt2, kim_2025corl_openvla} demonstrate that training on large and diverse robot datasets can improve generalization across tasks, while pretrained VLA models can be efficiently adapted to new settings through Low-Rank Adaptation~(LoRA)~\cite{hu2022lora} on downstream robot task data~\cite{kim_2025rss_openvlaoft, zheng2026xvla}. 
These advances reflect a growing emphasis on learning from diverse data to complement classical task- and platform-specific engineering. Pursuing this direction requires not only capable models but also practical ways to learn from data collected across distributed clients, tasks, and environments.
However, scaling VLA training by centralizing all robot data is often impractical. Robot data can be expensive to collect, large to store and transfer, and difficult to share across labs or institutions due to ownership or deployment constraints. Federated learning offers an appealing alternative: as illustrated in Fig.~\ref{fig:admm_vla_teaser}, each client trains on its local data while collaborating through a central server to update a shared global model, without uploading raw data~\cite{Mcmahan_fedavg_2017}. This setup is particularly well matched to robot learning, where data are naturally distributed across labs and task settings.

\begin{figure}[t]
    \centering
    \includegraphics[width=\columnwidth]{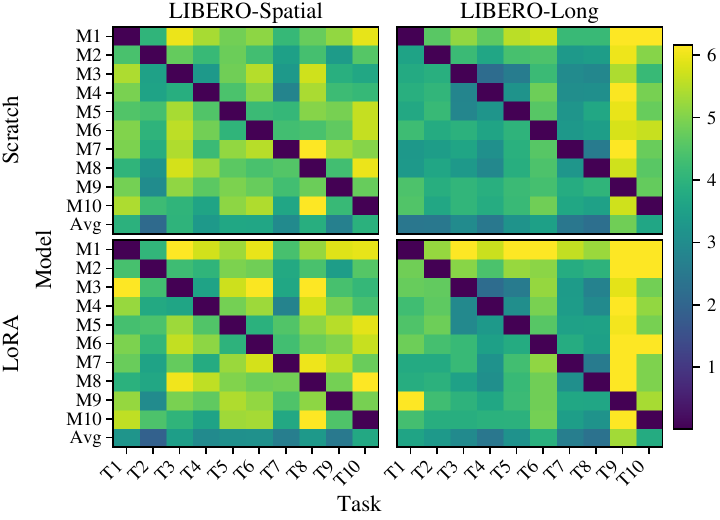}
    \caption{Cross-task training loss discrepancies of SmolVLA on LIBERO-Spatial and LIBERO-Long. }
    \label{fig:libero_loss_dis}
\end{figure}

Federated VLA training also introduces optimization challenges. Clients may observe different tasks, scenes, and action distributions, so local training can drift toward client-specific optima and degrade the global model under heterogeneously distributed data~\cite{Li_Yang_2020}. This challenge tends to be further amplified in federated LoRA fine-tuning, where adapter aggregation introduces additional instability under heterogeneous client data~\cite{sun2024improving, wang2024flora, bai2024flexlora}.
Fig.~\ref{fig:libero_loss_dis} illustrates this challenge on two LIBERO task suites for both training from scratch and LoRA fine-tuning. Each LIBERO suite contains 10 different tasks, so within each suite we independently train 10 SmolVLA~\cite{shukor2025smolvla} models, one per task, until convergence. We then evaluate each task-specialized model on the training sets of the other tasks from the same suite, producing a loss matrix. The resulting matrices show substantial cross-task discrepancies: a model that fits its own task well can incur much higher loss on other tasks. 
This suggests that averaging task-specialized client models may yield a poor global model.
 
To address these challenges, we present Co-VLA, an application of consensus ADMM~\cite{Boyd_2010} for federated VLA training and fine-tuning. Rather than relying only on parameter averaging, this algorithm explicitly constrains local client models to agree with a shared global model through consensus variables and dual updates. 
In contrast to prior federated VLA work~\cite{miao2025fedvla}, this formulation is model-agnostic: it applies to full-model VLA training from scratch, to LoRA-based fine-tuning by imposing consensus directly on the low-rank factors, and to sparse Low-Rank Adaptation~(SoRA), 
where trainable gates select useful rank components and inactive components can be pruned after training for a compact adapter~\cite{ding2023sparse}.
 
Our contributions are threefold. 
First, we formulate federated VLA training as a consensus problem and provide a practical synchronous ADMM instantiation for heterogeneous clients. 
Second, we show that the same consensus construction can be applied to LoRA and rank-adaptive SoRA, coupling the low-rank factors and rank gates across clients instead of aggregating independently drifting adapters. 
Third, we empirically show the effectiveness of Co-VLA with SmolVLA~\cite{shukor2025smolvla} and X-VLA~\cite{zheng2026xvla} on simulation and real-world data. Overall, Co-VLA outperforms representative federated baselines while achieving performance comparable to centralized training.

\section{Related work}
\label{sec:rw}
\subsection{Federated Learning}
Federated learning~\cite{Mcmahan_fedavg_2017} enables clients to collaboratively train a global model under the coordination of a central server without sharing raw data. The seminal algorithm FedAvg~\cite{Mcmahan_fedavg_2017} aggregates locally updated model parameters through weighted averaging; however, convergence can degrade significantly when the data is not independent and identically distributed (non-i.i.d.) across clients~\cite{Li_Yang_2020}. 
Subsequent work~\cite{reddi2021adaptive} improves FedAvg by replacing parameter averaging with more powerful server-side optimizers.
DiLoCo~\cite{douillard_2024want_diloco} shows that using Nesterov momentum~\cite{sutskever13nes}
as the outer optimizer achieves strong performance when fine-tuning large transformer models under non-i.i.d. client data distributions.


The alternating direction method of multipliers (ADMM)~\cite{Boyd_2010} provides a framework for consensus optimization, coupling local and global variables through quadratic penalties and dual updates.
In federated learning, ADMM-based approaches have been developed for different settings: FedPD~\cite{Zhang_FedPD_2021} adapts communication to data heterogeneity, while FedADMM~\cite{Zhou_FedADMM_2023, Wang_FedADMM_2022, Yi_Freris_2023} study inexact local updates and partial participation.
Related work has also explored asynchronous updates~\cite{Zhang_Kwok_2014, Bastianello_2021} and event-triggered communication~\cite{Er_2025}.
In this work, we empirically study consensus ADMM for federated VLA training and LoRA/SoRA fine-tuning under a simple setup with full client participation and synchronous communication: all clients perform local training and communicate their updates to the server at every round.

\subsection{Federated Low-Rank Adaptation} Low-Rank Adaptation~(LoRA)~\cite{hu2022lora} is one of the most widely used parameter-efficient fine-tuning methods and has been widely adopted for federated fine-tuning, including personalized federated learning~\cite{yi2024pfedlora, guo2025selective, zhang2026fedamole} and settings where clients have different computational resources~\cite{cho_2024emnlp_heterogeneous, bai2024flexlora, wang2024flora}.
In contrast, we assume clients have comparable resources and train a single global model from potentially non-i.i.d. client data, aiming to match centralized training.

Federated LoRA is more challenging than full-model fine-tuning because each client's model update is represented as the product of two low-rank matrices. Averaging these matrices separately and then multiplying them generally differs from averaging their products, creating an aggregation mismatch. 
SLoRA~\cite{babakniya2023slora} empirically shows that simple FedAvg over LoRA adapters can degrade under non-i.i.d.~data and proposes a data-driven initialization strategy to reduce this gap. Later methods analyze the aggregation mismatch and introduce different remedies. FFA-LoRA~\cite{sun2024improving} freezes one LoRA factor to stabilize aggregation, but this reduces the trainable capacity of the adapter. Other approaches, such as FLoRA~\cite{wang2024flora} and FlexLoRA~\cite{bai2024flexlora}, modify how LoRA updates are aggregated and redistributed by the server. 
Our method addresses the mismatch problem directly by the global consensus mechanism as detailed in Section~\ref{sec:federated_lora}.
Other federated LoRA methods study complementary directions, such as addressing rank capacity limits~\cite{yan2025federated} or sparsifying communication~\cite{kuo2024sparsity}.

\subsection{Vision-Language-Action Models}
VLAs extend vision-language models to robotic control by mapping visual observations and language instructions  to actions. Early works~\cite{octo_2023, google2023rt1} train policies from scratch on large-scale robot datasets, while more recent approaches~\cite{zitkovich23a_rt2, kim_2025corl_openvla, zheng2026xvla, shukor2025smolvla} build on pretrained vision-language models and continue training them on robot-specific data. Together, these methods show that diverse robotic data can improve both generalization and downstream performance. OpenVLA-OFT~\cite{kim_2025rss_openvlaoft} further demonstrates that pretrained VLA models can be adapted effectively to downstream robot tasks with LoRA. These advances make federated VLA training a promising direction: data from different robots, embodiments, and tasks could be used collaboratively to improve model performance without requiring all data to be centralized. However, federated learning for VLA remains underexplored. FedVLA~\cite{miao2025fedvla} is the first work that studies federated VLA training. It uses a Mixture-of-Experts~(MoE) VLA model that routes task-relevant inputs through adaptive experts on each client and aggregates client updates at the server according to expert activation similarity. Unlike FedVLA, which is tied to an MoE architecture, our ADMM-based approach is model-agnostic and applies to a broader range of VLAs.


\section{Method}
Federated VLA training must combine data from robot clients that may differ in tasks, environments, and action distributions, while not directly sharing each client's data to other clients. Simple parameter averaging can struggle in this setting because local updates may drift toward different task-specific optima under heterogeneous data. We therefore cast federated VLA training and fine-tuning as a global consensus optimization problem, where each robot optimizes a local model while ADMM explicitly encourages agreement with a shared global model through consensus constraints.

\label{sec:method}
\subsection{Federated VLA Training with Global Consensus}
We consider $N$ robot clients, where each robot client $i$ holds a local dataset $\mathcal{D}_i$. In the centralized setting, standard VLA training pools all data and minimizes the aggregate loss $\min_{\mathbf{x}\in\mathbb{R}^n}\sum_{i=1}^N f_i(\mathbf{x})$,
where $\mathbf{x}$ denotes the shared model parameters or LoRA factors and $f_i$ is the local action-supervision loss for dataset $\mathcal{D}_i$, such as cross-entropy~\cite{kim_2025corl_openvla}, $\ell_1$ regression~\cite{kim_2025rss_openvlaoft}, or diffusion loss~\cite{zheng2026xvla, shukor2025smolvla}.

For federated VLA training, we instead introduce a local copy $\mathbf{x}_i$ of the model parameters for each robot and a global consensus variable $\mathbf{z}$ maintained by the central server. This yields the following global consensus formulation~\cite{Boyd_2010}:
\begin{equation}
    \min_{\mathbf{x}_1, \ldots, \mathbf{x}_N,\, \mathbf{z}}~ \sum_{i=1}^N f_i(\mathbf{x}_i) \quad  \text{s.t.} \; \mathbf{x}_i = \mathbf{z}, \; i = 1, \ldots, N,
\label{eqn:consensus}
\end{equation} 
where the constraints $\mathbf{x}_i = \mathbf{z}$ enforce local--global model agreement despite heterogeneous client data.

\begin{algorithm}[tb]
\caption{Distributed learning with over-relaxed ADMM}
\label{alg:over_relaxed_consensus}
\begin{algorithmic}
\REQUIRE Local objective $f_i$, hyper-parameters $\rho$, $\alpha$\\
\STATE \textbf{Initialize:} $\mathbf{x}_i^0=\widehat{\mathbf{x}}_i^0=\mathbf{x}^0$, $\mathbf{z}^0=\mathbf{x}^0$, $\mathbf{u}_i^{-1}=\mathbf{0}$
\FOR{$k=0$ to $t_{\mathrm{max}}$}
\FOR{$i=1$ to $N$} 
\hspace*{-\fboxsep}{\colorbox{YellowGreen!50}{\parbox{\dimexpr\linewidth-2\fboxsep\relax}{
\STATE $\mathbf{u}_i^k= \mathbf{u}_i^{k-1}+ \widehat{\mathbf{x}}_i^k - \mathbf{z}^k$   \hfill \COMMENT{Client $i$}
\STATE $\mathbf{x}_i^{k+1} = \arg \min_{\mathbf{x}_i} f_i(\mathbf{x}_i)+\frac{\rho}{2}\|\mathbf{x}_i-\mathbf{z}^k+\mathbf{u}_i^k\|_2^2$
\STATE $\widehat{\mathbf{x}}_i^{k+1}= \alpha \mathbf{x}_i^{k+1} + (1-\alpha) \mathbf{z}^{k}$
}}}\ENDFOR

\hspace*{-\fboxsep}{\colorbox{LightGray}{\parbox{\dimexpr\linewidth-2\fboxsep\relax}{
\STATE $\mathbf{z}_{k+1}=\frac{1}{N}\sum_i^N \left( \widehat{\mathbf{x}}_i^{k+1}+\mathbf{u}_i^k\right)$ \hfill \COMMENT{Server} 
}}}
\ENDFOR 
\end{algorithmic}
\end{algorithm}

ADMM~\cite{Boyd_2010} solves~\eqref{eqn:consensus} by alternating between a local minimization step on each client and a global aggregation step on the central server (see Algorithm~\ref{alg:over_relaxed_consensus}). In each round, the client $i$ minimizes its local objective augmented by a quadratic proximity term that pushes the update towards the current global model; in practice, the minimization to calculate local models is replaced by a fixed number of local gradient descent steps. The parameter $\rho>0$ controls the strength of this coupling, and the dual variable $\mathbf{u}_i$ is a Lagrange multiplier that accumulates the consensus gap ($\widehat{\mathbf{x}}_i-\mathbf{z}$) across rounds. 
The server then averages the client updates to calculate the new global model, i.e.,~consensus variable.  We adopt an over-relaxed variant of ADMM with relaxation parameter $\alpha \in (0, 2)$, which can accelerate convergence relative to standard ADMM ($\alpha = 1$)~\cite{Nishihara_2015, Er_2025}. 

ADMM only introduces modest additional computation over FedAvg.
On the client side, each round adds a dual-variable update and a quadratic regularization term in the primal update. The server averages the client-side primal-dual sums $\widehat{\mathbf{x}}_i + \mathbf{u}_i$ and broadcasts the updated global parameter $\mathbf{z}$ to clients, as illustrated in Fig.~\ref{fig:admm_vla_teaser}. Thus, communication bandwidth remains on the same order as FedAvg.

\subsection{Federated LoRA with Global Consensus}
\label{sec:federated_lora}


LoRA fine-tunes a pretrained weight matrix $\mathbf{W}_0 \in \mathbb{R}^{d\times k}$ by learning a low-rank update instead of modifying all parameters directly. Specifically, the adaptation is written as $\mathbf{W}\mathbf{s}=(\mathbf{W}_0+\Delta\mathbf{W})\mathbf{s}$ with $\Delta\mathbf{W}=\mathbf{B}\mathbf{A}$, where $\mathbf{A} \in \mathbb{R}^{r\times k}$ and $\mathbf{B} \in \mathbb{R}^{d\times r}$ are trainable low-rank factors with $r \ll \min(d,k)$, and $\mathbf{s} \in \mathbb{R}^k$ denotes the input state. In federated LoRA, each client $i$ maintains local factors $\mathbf{A}_i$ and~$\mathbf{B}_i$, which together define its local update $\Delta \mathbf{W}_i=\mathbf{B}_i\mathbf{A}_i$.

As mentioned in the related work, LoRA with FedAvg is particularly sensitive to heterogeneous client data because the server typically aggregates the low-rank factors separately. Let $\mathbb{E}$ denote averaging over clients $i\in\{1,\ldots,N\}$. For client-specific factors $\mathbf{A}_i$ and $\mathbf{B}_i$, averaging the factors does not in general recover the average update, i.e., $\mathbb{E}[\Delta \mathbf{W}_i]=\mathbb{E}[\mathbf{B}_i\mathbf{A}_i]\neq\mathbb{E}[\mathbf{B}_i]\,\mathbb{E}[\mathbf{A}_i]$, 
which can make naive FedAvg over LoRA adapters unreliable under non-i.i.d.~data~\cite{sun2024improving, wang2024flora, bai2024flexlora}. Existing methods address this mismatch with additional mechanisms: FLoRA~\cite{wang2024flora} collects and stacks the factor in the server and redistributes stacked factors across clients, which increases communication with the number of clients; and FlexLoRA~\cite{bai2024flexlora} reconstructs full updates on the server and refactorizes them with SVD, introducing heavy computation and approximation error.

Our global consensus formulation in Eq.~\eqref{eqn:consensus} addresses the mismatch more directly. Instead of averaging independently drifting LoRA factors, ADMM encourages all clients to agree on shared global factors by enforcing
$\mathbf{A}_i \approx \mathbf{A}$ and $\mathbf{B}_i \approx \mathbf{B}$ across clients. 
This allows federated LoRA training to retain the communication efficiency of low-rank adapters while reducing the bilinear aggregation mismatch caused by heterogeneous local updates.

\subsection{Federated Sparse LoRA Based on ADMM}
SoRA~\cite{ding2023sparse} augments LoRA with a trainable gate that selects useful rank components during fine-tuning. For an input feature $\mathbf{s}$, the LoRA update is written as $\Delta \mathbf{W}\mathbf{s}
    = \mathbf{B}\bigl(\mathbf{g} \odot (\mathbf{A}\mathbf{s})\bigr)$,
where $\mathbf{g}\in\mathbb{R}^r$ gates the $r$ rank components and $\odot$ denotes element-wise multiplication. Starting from the prescribed initial rank $r$, SoRA automatically adapts the effective rank by adding an $\ell_1$ penalty on $\mathbf{g}$ and solving the resulting objective with proximal gradient updates rather than ADMM.  After training, removing zero-gated rows of $\mathbf{A}$ and columns of $\mathbf{B}$ yields a compact LoRA module with an automatically selected rank.

This idea fits naturally into our consensus formulation by including the gate in the federated LoRA variables. Let $\mathbf{x}_i=\{\mathbf{A}_i,\mathbf{B}_i,\mathbf{g}_i\}$ be the client variables and $\mathbf{z}=\{\mathbf{A},\mathbf{B},\mathbf{g}\}$ be the global variables. We solve
\squeezeeq{
\begin{equation}
    \min_{\mathbf{x}_1,\ldots,\mathbf{x}_N,\mathbf{z}}~
     \sum_{i=1}^N f_i(\mathbf{x}_i) + \lambda\|\mathbf{g}\|_1  \quad  \text{s.t.~~} \; \mathbf{x}_i = \mathbf{z}, \; i = 1, \ldots, N,
\label{eqn:l1_consensus}
\end{equation}}
where $\lambda$ controls the sparsity of the shared gate. Since the $\ell_1$ norm is placed only on the global gate, the local client update remains the same as in Algorithm~\ref{alg:over_relaxed_consensus}. The server update for $\mathbf{A}$ and $\mathbf{B}$ remains an averaging step, while the gate is updated by the soft threshold operation~\cite{Boyd_2010}:
\begin{equation}
    \mathbf{g}^{k+1}
    = S_{\lambda/(N\rho)}\left(\frac{1}{N}\sum_{i=1}^N
    \left(\widehat{\mathbf{g}}_i^{k+1}+\mathbf{u}_{g,i}^k\right)\right),
\label{eqn:gate_update}
\end{equation}
where the soft threshold operation is defined element-wise as
\begin{equation}
    S_{\xi}(v_j)
    =
    \begin{cases}
        v_j - \xi, & v_j > \xi, \\
        0, & |v_j| \leq \xi, \\
        v_j + \xi, & v_j < -\xi,
    \end{cases}
    \quad j=1,\ldots,r .
\label{eqn:soft_threshold_gate}
\end{equation}
Intuitively, soft thresholding shrinks each gate value toward zero and sets entries within the threshold to exactly zero, thereby inducing sparsity in the rank components.
Federated sparse LoRA adds only server-side soft-thresholding, preserving client-side training.


\section{Experiments}
\label{sec:result}
We evaluate Co-VLA with SmolVLA~\cite{shukor2025smolvla} and X-VLA~\cite{zheng2026xvla} using their LeRobot implementations~\cite{cadene2024lerobot}. SmolVLA is a compact 0.45B-parameter VLA with a frozen pretrained VLM backbone and a flow-matching action expert. We use SmolVLA to study full-model training and LoRA/SoRA fine-tuning on LIBERO, as well as full-model training on real-world data. We also evaluate LoRA and SoRA fine-tuning on LIBERO with X-VLA, a larger 0.9B-parameter flow-matching VLA whose learnable soft prompts condition its transformer policy.
All models use \texttt{bf16} unless otherwise stated.

\subsection{Evaluation with Simulated Data}
\subsubsection{Centralized training baseline}
For SmolVLA, we train the centralized baselines using the learning rates, batch sizes, and training budgets following the upstream LeRobot implementation~\cite{cadene2024lerobot}. For training from scratch, we use a batch size of 64, a learning rate of $10^{-4}$, and 100,000 optimization steps on a single H100 GPU. For LoRA fine-tuning, we use a batch size of 32, rank 64, a learning rate of $10^{-3}$, and 100,000 optimization steps on a single H100 GPU, applying LoRA to all linear layers. The default LeRobot configuration applies LoRA only to the \texttt{q} and \texttt{v} projections and the state-action projection matrices, empirically we found that extending LoRA to all linear layers improves performance.

For X-VLA, we use the LeRobot implementation and use the hyperparameters described in the original paper. We train the centralized baseline on two H100 GPUs with a per-GPU batch size of 64. We apply rank-8 LoRA to all linear layers, including those in the VLM, and additionally tune the soft prompt, action encoder, and action decoder. The base learning rate is $10^{-4}$, with reduced learning rates for the soft prompt and VLM, and training runs for 40,000 steps.

\subsubsection{LIBERO benchmark}
LIBERO~\cite{liu2023libero} contains four task suites with 10 tasks each, where each task is a distinct language-conditioned manipulation problem with multiple demonstration episodes for training, such as picking an object from a specified location or placing an object into a target container. LIBERO-Spatial tests spatial reasoning with the same objects arranged in different layouts; LIBERO-Object varies the manipulated objects under similar scene layouts; LIBERO-Goal varies the target manipulation objective while keeping objects and layouts similar; and LIBERO-Long contains long-horizon tasks with more diverse objects, layouts, and goals. We report average success rate, computed over 500 evaluation trials per task suite, corresponding to 50 episodes for each of the 10 tasks.

\subsubsection{Federated training setup}
For each LIBERO suite, we simulate a 10-client federated setting, assigning one task to each client. Local training hyperparameters, including batch size, learning rate, and LoRA rank, match the centralized setup, and each client performs 100 local steps per communication round. We run the experiments on a single node with four H100 GPUs, assigning two or three clients to each GPU and colocating the server with one GPU process.
We compare Co-VLA with two general-purpose federated baselines, FedAvg and DiLoCo. We select the Co-VLA and DiLoCo hyperparameters through a grid search on the LIBERO-Long training set. For Co-VLA, we use $\rho=10^{-4}$ and an over-relaxation parameter of $\alpha=1.8$; for DiLoCo, we use an outer learning rate of 0.7 and momentum of 0.9. For LoRA fine-tuning, we additionally compare with two federated LoRA methods, FLoRA and FlexLoRA.

Table~\ref{tab:qualitative_comparison} summarizes the qualitative differences between Co-VLA and representative federated methods. The key distinction is that Co-VLA is a flexible algorithm that supports various VLA architectures, LoRA-aware aggregation, and sparse rank adaptation under the same global consensus view, and the global consensus formulation naturally supports non-i.i.d. data across clients.

\begin{table}[tb]
    \centering
    \caption{Qualitative comparison of federated training strategies.}
    \label{tab:qualitative_comparison}
    \setlength{\tabcolsep}{4pt}
    \footnotesize
    \begin{tabular}{@{}lcccc@{}}
        \toprule
        \textbf{Method} & \textbf{General VLA} & \textbf{LoRA-aware} & \textbf{SoRA} & \textbf{Non-i.i.d.} \\
        \midrule
        FedAvg & \cmark & \xmark & \xmark & \xmark \\
        DiLoCo & \cmark & \xmark & \xmark & \cmark \\
        FLoRA & \xmark & \cmark & \xmark & \cmark \\
        FlexLoRA & \xmark & \cmark & \xmark & \cmark \\
        FedVLA & \xmark & \xmark & \xmark & \cmark \\
        Co-VLA & \cmark & \cmark & \cmark & \cmark \\
        \bottomrule
    \end{tabular}
\end{table}

\subsubsection{Results of training from scratch for SmolVLA}
Table~\ref{tab:smolvla_scratch} compares Co-VLA with FedAvg and DiLoCo under the same communication budget of 1,500 rounds, corresponding to 150,000 local optimization steps per client. Relative to the 100,000-step centralized baseline, the federated methods therefore perform $1.5\times$ as many optimization steps per client and  process $10\times$ as many training samples per step across the ten clients.
Co-VLA closely matches centralized training across the LIBERO suites and achieves the best federated performance on Spatial and Goal. DiLoCo performs competitively and even outperforms all other methods on LIBERO-Long, while FedAvg generally underperforms the other methods.

\begin{table}[tb] 
    \centering
    \caption{Success rate on LIBERO task suites of SmolVLA training from scratch. Co-VLA achieves performance comparable to DiLoCo and centralized training while outperforming FedAvg.}
    \label{tab:smolvla_scratch}
    \begin{tabular}{lcccc}
        \toprule
        \textbf{Method} & \textbf{Spatial} & \textbf{Object} & \textbf{Goal} & \textbf{Long}\\
        \midrule
        Co-VLA   & \textbf{0.80} & {0.91} & \textbf{0.92} & 0.71 \\
        FedAvg & 0.76 & 0.89 & 0.91 & 0.63  \\
        DiLoCo & 0.77 & \textbf{0.93} & 0.90 & \textbf{0.76}  \\
        \midrule
        {Centralized training} & 0.81 & 0.95 & 0.94 & 0.70 \\
        \bottomrule
    \end{tabular}
\end{table}

\begin{figure}[tb]
    \centering
    \begin{subfigure}{0.49\linewidth}
        \centering
        \includegraphics[width=\linewidth]{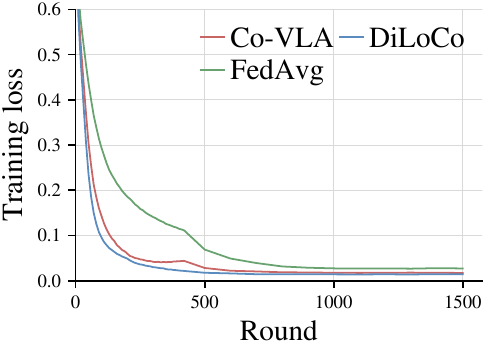}
        \caption{LIBERO-Spatial}
    \end{subfigure}
    \hfill
    \begin{subfigure}{0.49\linewidth}
        \centering
        \includegraphics[width=\linewidth]{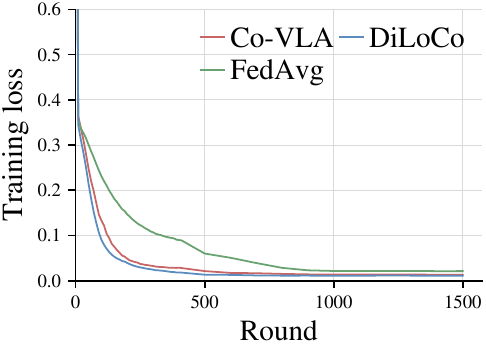}
        \caption{LIBERO-Object}
    \end{subfigure}
    \par\vspace{1ex}
    \begin{subfigure}{0.49\linewidth}
        \centering
        \includegraphics[width=\linewidth]{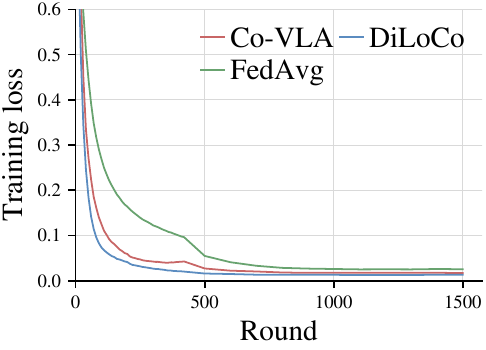}
        \caption{LIBERO-Goal}
    \end{subfigure}
    \hfill
    \begin{subfigure}{0.49\linewidth}
        \centering
        \includegraphics[width=\linewidth]{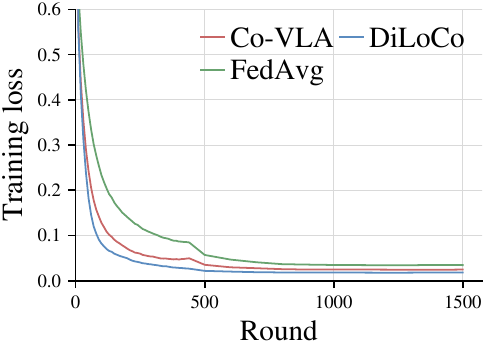}
        \caption{LIBERO-Long}
    \end{subfigure}
    \caption{Loss curves for SmolVLA training from scratch across LIBERO task suites.}
    \label{fig:loss_curve_scratch}
\end{figure}

Figure~\ref{fig:loss_curve_scratch} shows the training loss curves, where the flow-matching loss is periodically evaluated using the global model on the full training set. These curves reflect the convergence behavior of different methods. DiLoCo converges the fastest for training from scratch, Co-VLA achieves comparable convergence speed, while FedAvg converges more slowly than the other two methods.
All methods are trained until the loss plateaus with a relative loss change smaller than 5\%.

\subsubsection{Results of LoRA fine-tuning for SmolVLA and X-VLA}
Table~\ref{tab:smolvla_lora} reports the SmolVLA LoRA fine-tuning results over 2,500 communication rounds and includes FLoRA and FlexLoRA as additional federated LoRA baselines. Following FlexLoRA, we compute adapter redistribution with \texttt{torch.svd}. Because training uses \texttt{bf16}, we cast the parameters involved in the SVD computation to \texttt{float32} and convert them back afterward. For Co-VLA SoRA, we keep the gate parameters in \texttt{float32}, as using \texttt{bf16} caused them to collapse to zero. We use $\lambda=10^{-5}$, and the percentages in parentheses indicate the fraction of adapter parameters retained after pruning.

Co-VLA LoRA achieves the strongest performance on Spatial and Object and remains competitive on Goal and Long. Co-VLA SoRA matches or slightly improves upon Co-VLA LoRA on Goal and Long while retaining only about 55--60\% of the adapter parameters, although its performance drops moderately on Spatial. FLoRA and FlexLoRA perform similarly and generally outperform FedAvg, whereas DiLoCo underperforms in the LoRA setting. Federated LoRA performance varies relatively little on LIBERO-Long.
\begin{table}[tb]
    \centering
    \caption{Success rate on LIBERO task suites of SmolVLA LoRA fine-tuning. Our Co-VLA variants outperform other LoRA-based baselines across most task suites.}
    \label{tab:smolvla_lora}
    \setlength{\tabcolsep}{3pt}
    \begin{tabular}{@{}lcccc@{}}
        \toprule
        \textbf{Method} & \textbf{Spatial} & \textbf{Object} & \textbf{Goal} & \textbf{Long} \\
        \midrule
        Co-VLA~LoRA   & \textbf{0.75} & \textbf{0.81} & {0.87} & 0.51 \\
        Co-VLA~SoRA   & {0.69} (60\%) & 0.79 (59\%) & \textbf{0.88} (55\%)  & 0.53 (58\%)  \\
        FedAvg~LoRA & 0.61 & 	0.67 & 0.79 & {0.52}  \\
        DiLoCo~LoRA & 0.47 & 0.61 & 0.77 & 0.51  \\
        FLoRA  & 0.66 & 0.79 & 0.81 & 0.50 \\
        FLoRA (7500) & 0.71 & 0.80 & 0.84 & \textbf{0.58} \\
        FlexLoRA & 0.66 & 0.77 & 0.81 & {0.54} \\
        \midrule
        {Centralized training} & 0.76 & 0.80 & 0.84 & 0.57  \\
        \bottomrule
    \end{tabular}
\end{table}

\begin{figure}[tb]
    \centering
    \begin{subfigure}{0.49\linewidth}
        \centering
        \includegraphics[width=\linewidth]{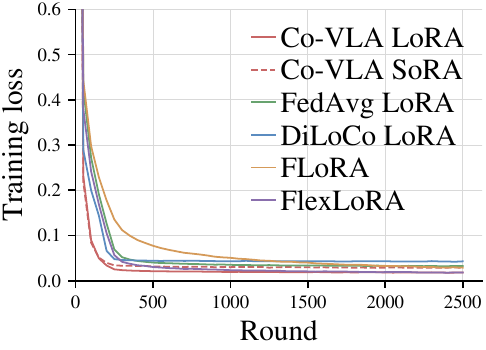}
        \caption{LIBERO-Spatial}
    \end{subfigure}
    \hfill
    \begin{subfigure}{0.49\linewidth}
        \centering
        \includegraphics[width=\linewidth]{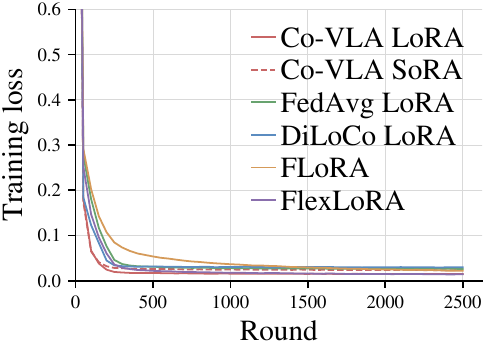}
        \caption{LIBERO-Object}
    \end{subfigure}
    \par\vspace{1ex}
    \begin{subfigure}{0.49\linewidth}
        \centering
        \includegraphics[width=\linewidth]{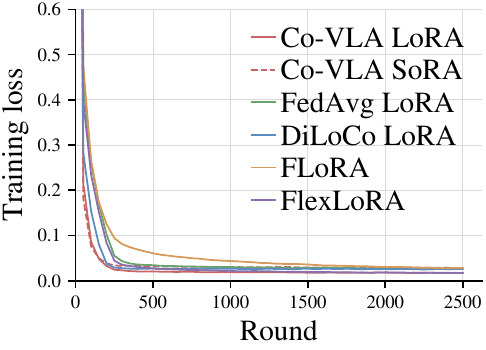}
        \caption{LIBERO-Goal}
    \end{subfigure}
    \hfill
    \begin{subfigure}{0.49\linewidth}
        \centering
        \includegraphics[width=\linewidth]{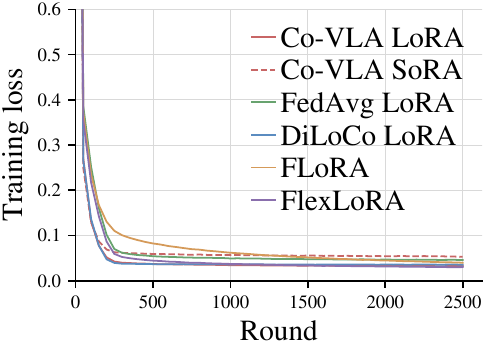}
        \caption{LIBERO-Long}
    \end{subfigure}
    \caption{Loss curves for SmolVLA LoRA fine-tuning across LIBERO task suites.}
    \label{fig:loss_curve_lora}
\end{figure}

Figure~\ref{fig:loss_curve_lora} shows the training loss curves for LoRA fine-tuning, computed as described above. Co-VLA generally converges faster than the other methods. By 2,500 communication rounds, all methods except FLoRA meet the loss-plateau criterion of a relative loss change below 5\%. We therefore also report FLoRA's performance at 7,500 rounds, by which point it meets the same criterion. The extended training improves FLoRA's performance, particularly on LIBERO-Long, but triples its communication cost.

We also compare the per-round training time of FedAvg LoRA, Co-VLA LoRA, and the SoRA variant. FedAvg LoRA takes $81.22 \pm 7.87$ seconds per round, while Co-VLA LoRA takes $81.70 \pm 7.82$ seconds, suggesting only modest ADMM overhead. The SoRA variant takes $85.27 \pm 8.07$ seconds per round, mainly due to its additional gate parameters.

Table~\ref{tab:xvla_lora} reports the X-VLA LoRA fine-tuning results after 900 communication rounds. Co-VLA LoRA performs comparably to centralized training across the LIBERO task suites and substantially outperforms FlexLoRA. Co-VLA SoRA achieves slightly lower success rates while retaining about 83--89\% of the LoRA parameters. The performance gap between Co-VLA LoRA and SoRA may result from the small initial LoRA rank of 8, as adaptive-rank methods such as SoRA typically benefit from a larger starting rank. 
By 900 communication rounds, training loss has converged for both Co-VLA variants but not for FlexLoRA. We therefore also report FlexLoRA's results at 2,500 rounds, by which point it meets the loss-plateau criterion of a relative loss change below 5\%. Despite this extended training, FlexLoRA still underperforms Co-VLA.
Under the same communication budget, FLoRA has not yet met the loss-plateau criterion and performs substantially worse than the other methods.

\begin{table}[tb]
    \centering
    \caption{Success rate on LIBERO task suites for X-VLA LoRA fine-tuning. Percentages in parentheses indicate the retained LoRA parameters after SoRA pruning.}
    \label{tab:xvla_lora}
    \setlength{\tabcolsep}{3pt}
    \begin{tabular}{@{}lcccc@{}}
        \toprule
        \textbf{Method} & \textbf{Spatial} & \textbf{Object} & \textbf{Goal} & \textbf{Long} \\
        \midrule
        Co-VLA~LoRA   & \textbf{0.88} & \textbf{0.97} & \textbf{0.93} & \textbf{0.78}\\
        Co-VLA~SoRA   & 0.83 (86\%) & 0.91 (83\%) & {0.85} (86\%)  & 0.70 (89\%) \\
        FlexLoRA & 0.64 & 0.62 & 0.55 & 0.54 \\
        FlexLoRA (2500) & 0.74 & 0.70 & 0.64 & 0.54 \\

        \midrule
        {Centralized training} & 0.87 & 0.98 & 0.93 & 0.74\\
        \bottomrule
    \end{tabular}
\end{table}

\subsection{Evaluation with Real-World Data for SmolVLA}
\begin{figure*}[tb]
    \centering
    \newcommand{\gridpanel}[1]{\includegraphics[width=0.31\linewidth]{#1}}

    \begin{subfigure}{0.496\textwidth}
        \centering
        \gridpanel{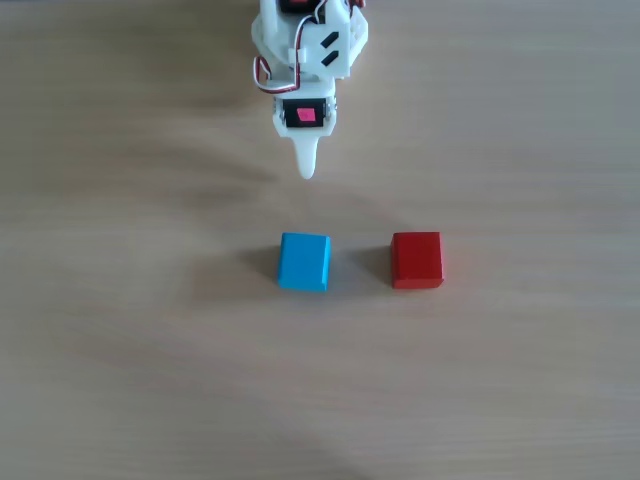}
        \gridpanel{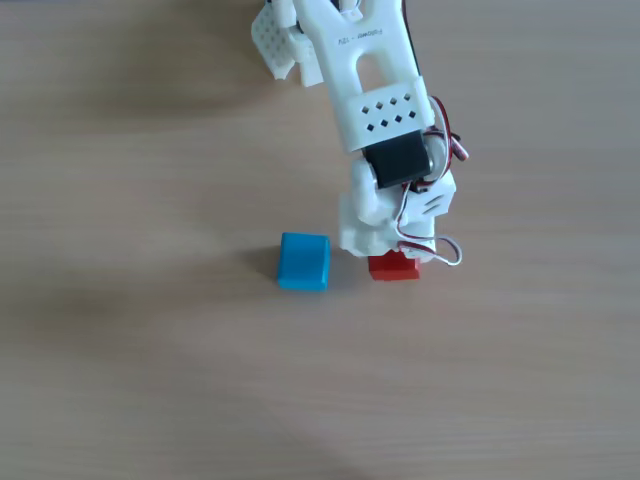}
        \gridpanel{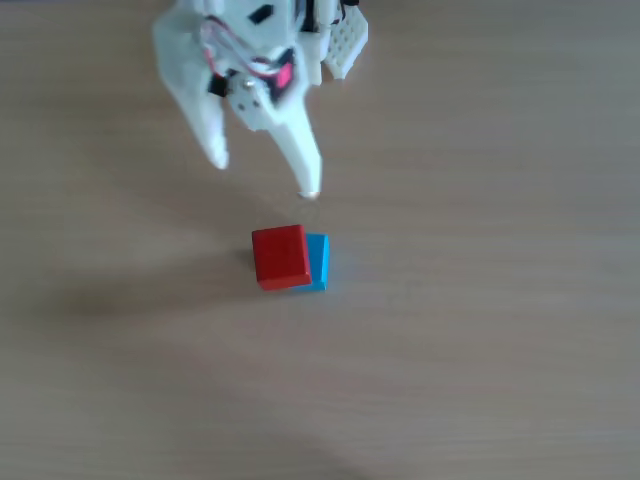}\\[0.5ex]
        \gridpanel{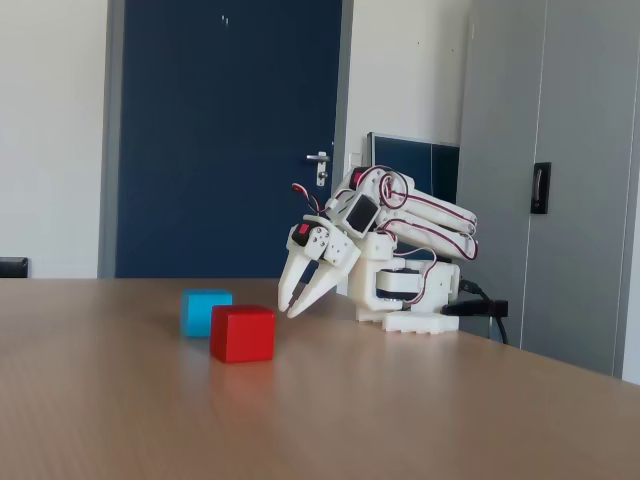}
        \gridpanel{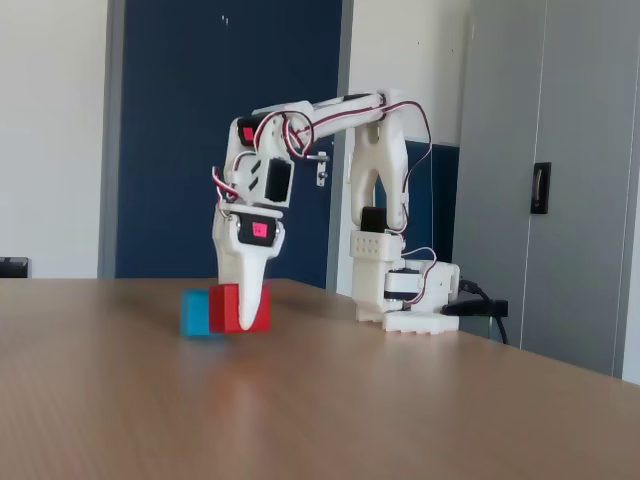}
        \gridpanel{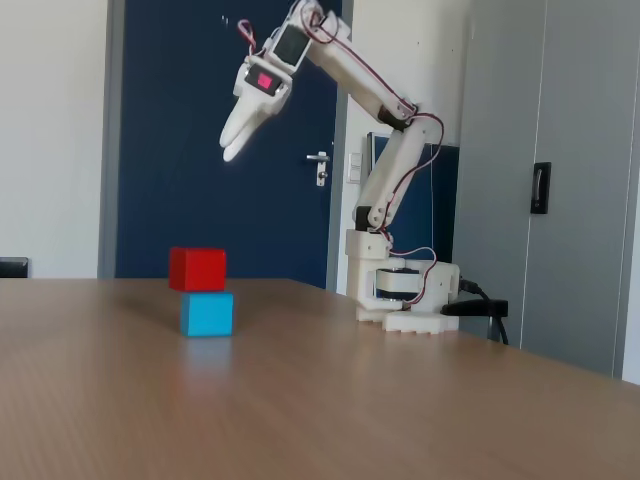}
        \caption{Stack}
        \label{fig:stack}
    \end{subfigure}\hspace{0.006\textwidth}%
    \begin{subfigure}{0.496\textwidth}
        \centering
        \gridpanel{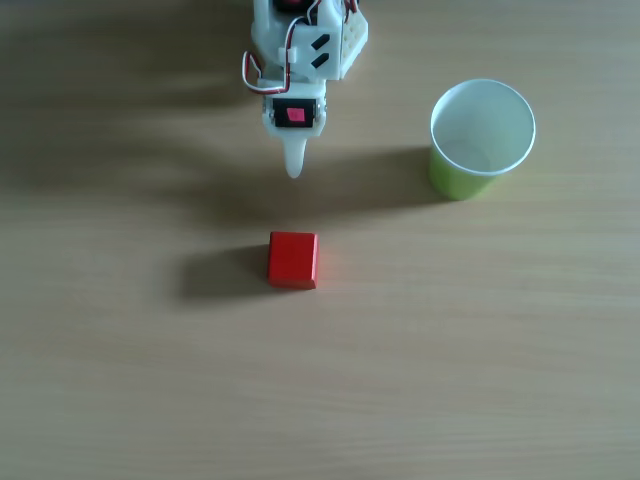}
        \gridpanel{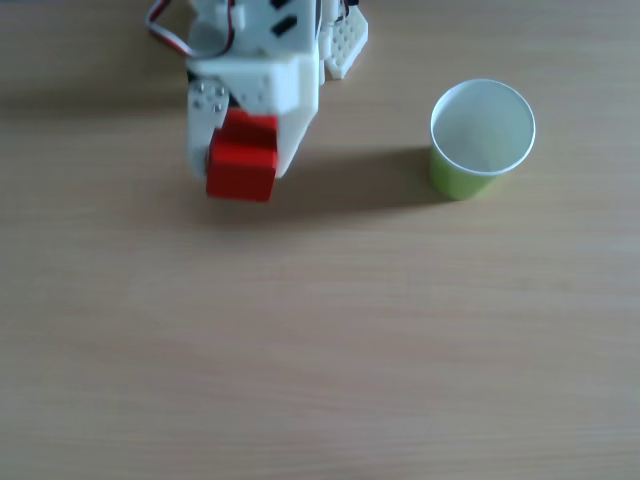}
        \gridpanel{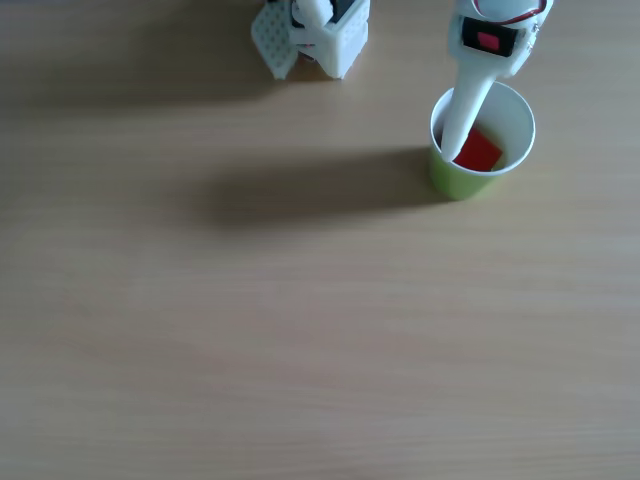}\\[0.5ex]
        \gridpanel{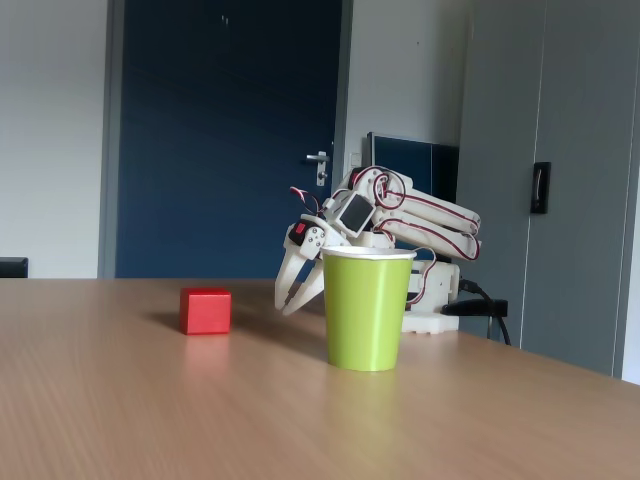}
        \gridpanel{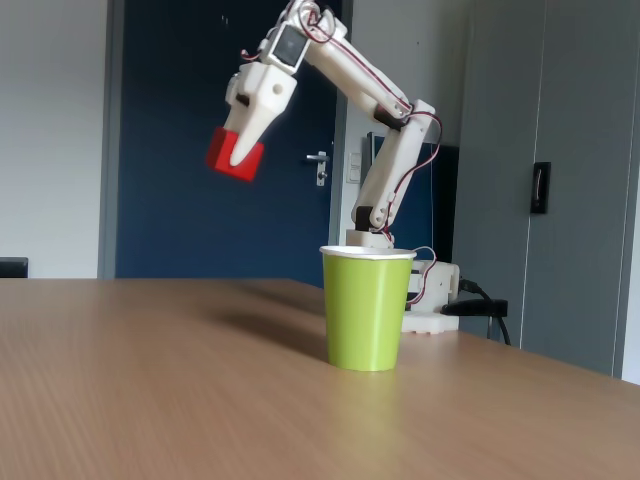}
        \gridpanel{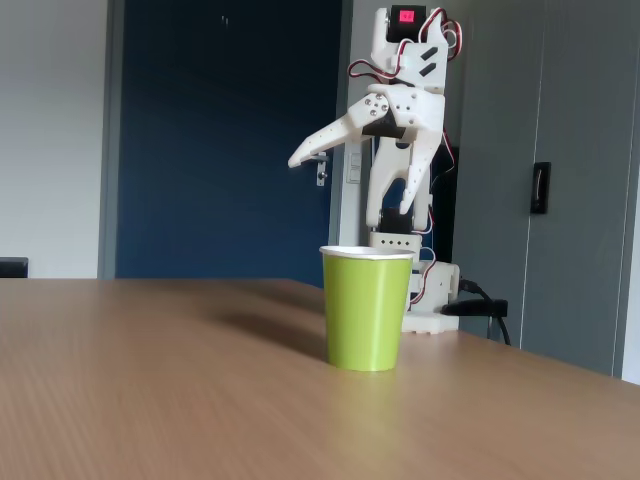}
        \caption{Pick and place}
        \label{fig:pick}
    \end{subfigure}

    \par\vspace{1ex}
    \begin{subfigure}{0.496\textwidth}
        \centering
        \gridpanel{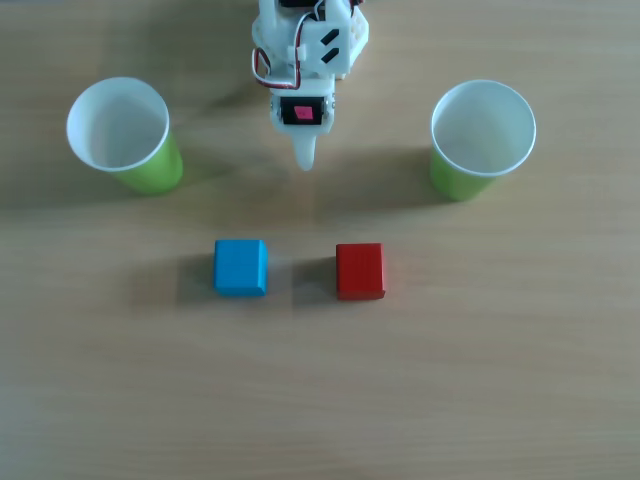}
        \gridpanel{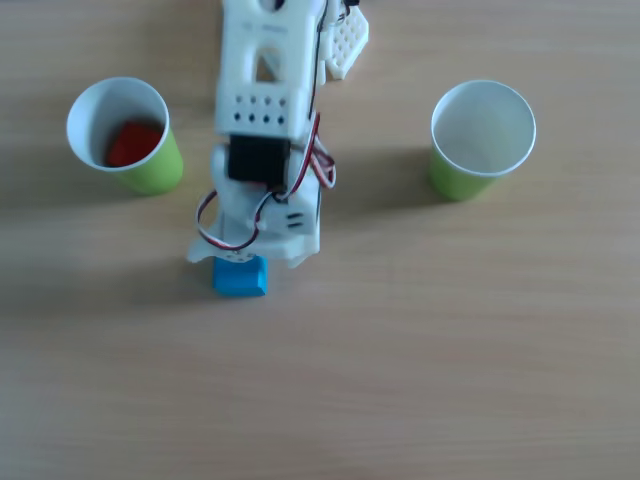}
        \gridpanel{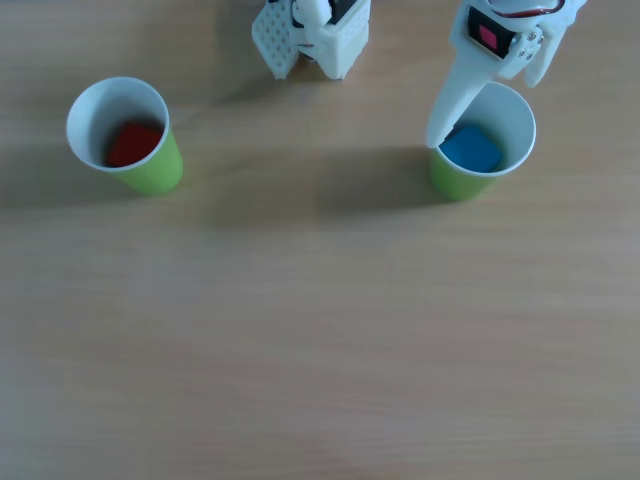}\\[0.5ex]
        \gridpanel{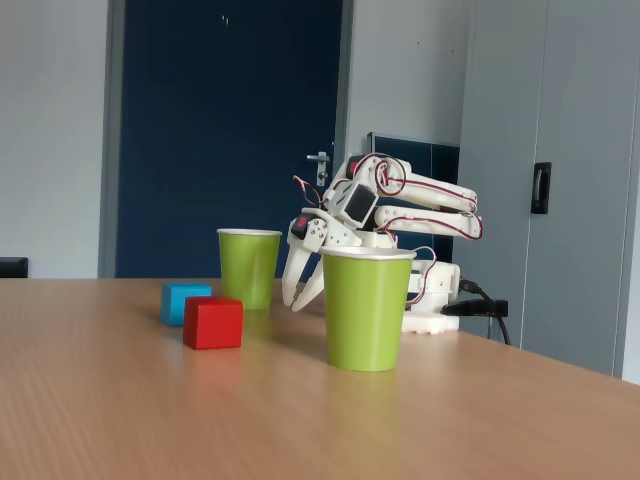}
        \gridpanel{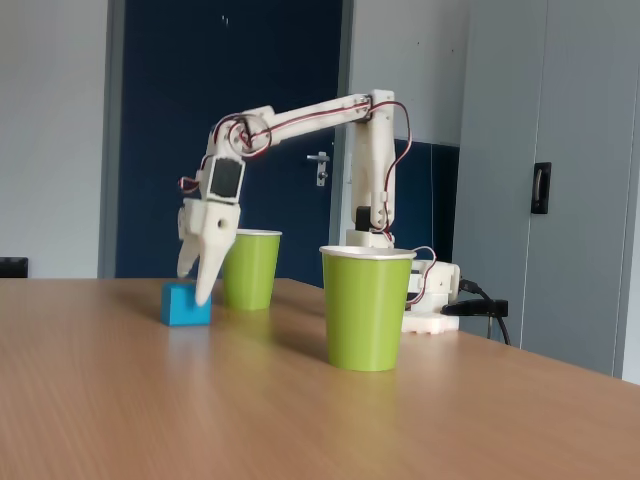}
        \gridpanel{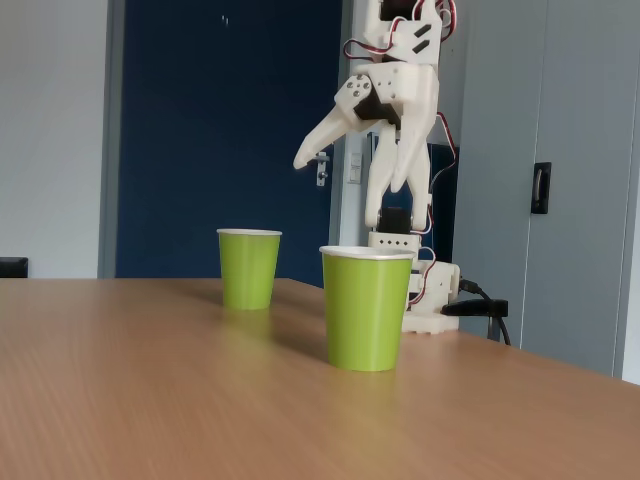}
        \caption{Sort}
        \label{fig:sort}
    \end{subfigure}\hspace{0.006\textwidth}%
    \begin{subfigure}{0.496\textwidth}
        \centering
        \gridpanel{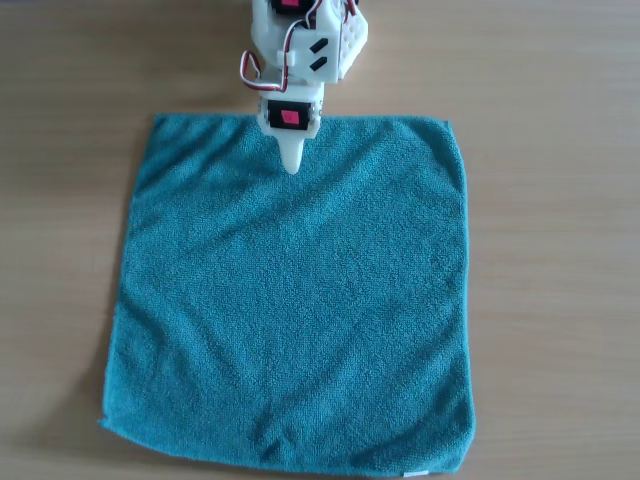}
        \gridpanel{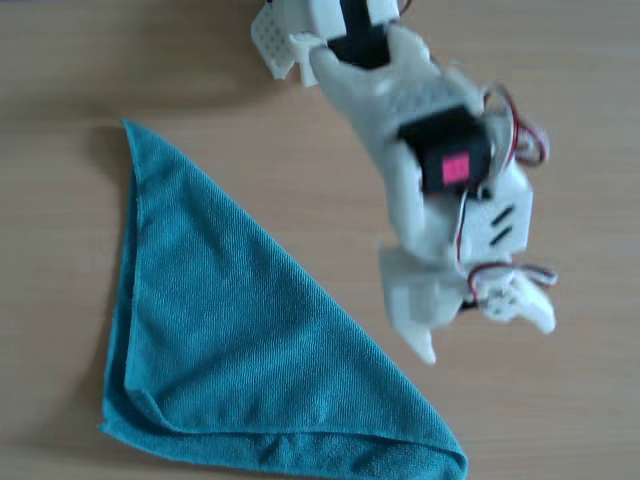}
        \gridpanel{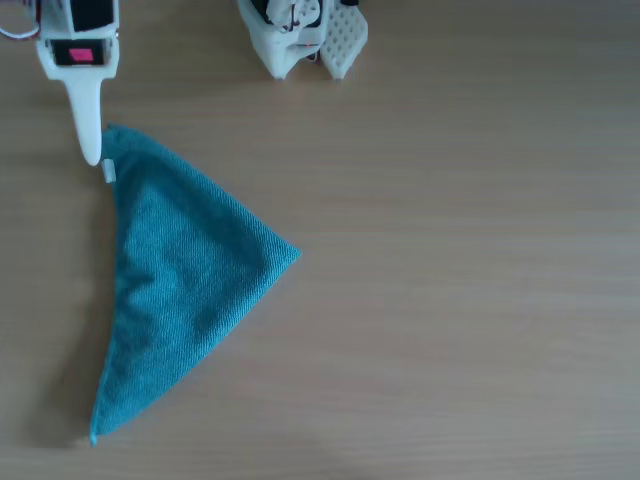}\\[0.5ex]
        \gridpanel{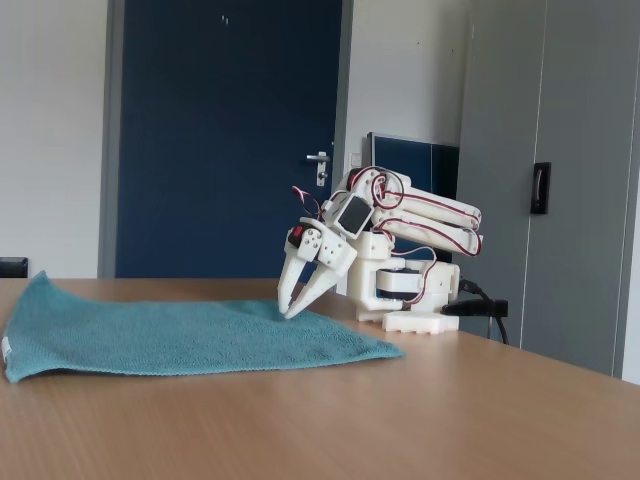}
        \gridpanel{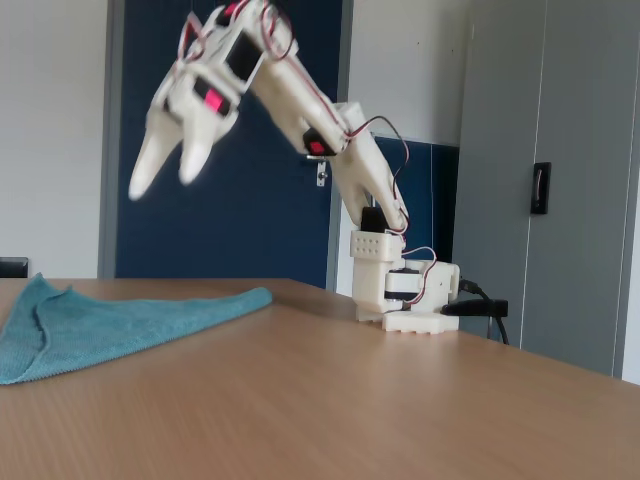}
        \gridpanel{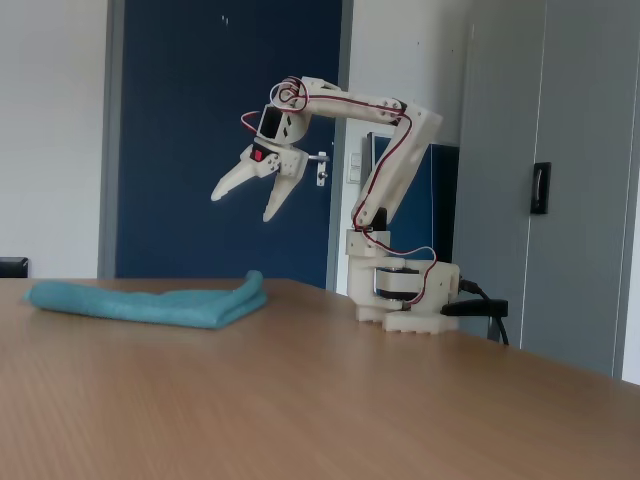}
        \caption{Fold}
        \label{fig:fold}
    \end{subfigure}

    \caption{Real-world closed-loop experiments.}
    \label{fig:realworld_exp}
\end{figure*}
\subsubsection{Open-loop evaluation}
We construct a mixed real-world dataset by selecting four datasets from the Open X-Embodiment collection~\cite{open_x_embodiment_rt_x_2023}: Berkeley Bridge~\cite{walke2023bridgedata}, FMB~\cite{luo2024fmb}, Jaco Play~\cite{dass2023jacoplay}, and Fractal~\cite{google2023rt1}. We use their LeRobot-format versions~\cite{cadene2024lerobot}. In all selected datasets, actions are represented in end-effector pose space.
For each dataset, we first randomly reserve 100 episodes for testing and then sample a subset of the remaining episodes for training, resulting in a total of 402,753 training transitions and the imbalanced data distribution summarized in Table~\ref{tab:real_world_dataset_settings}. We use this imbalanced setting to reflect a realistic multi-source robot learning scenario, where data collected across platforms and institutions naturally differ in scale and episode length.
The mixed dataset is highly heterogeneous in robot embodiments, scenes, visual inputs, and camera configurations.

As a centralized baseline, we train SmolVLA on the combined dataset for 100,000 steps using four H100 GPUs with a per GPU batch size of 128.
In the federated setting, we use four clients, each containing data from a single source dataset, and use the batch size of 128 for each client. We train all federated methods for 1,000 communication rounds with 100 local steps per round, by which point all meet the same loss-plateau criterion as in the preceding experiments. Co-VLA and DiLoCo use the same hyperparameters as in the LIBERO experiments.

\begin{table}[tb]
    \centering
    \caption{Dataset settings for mixed real-world dataset.}
    \label{tab:real_world_dataset_settings}
    \begin{tabular}{llcc}
        \toprule
        \textbf{Dataset} & \textbf{Robot type} & \textbf{Action space} & \textbf{Training share} \\
        \midrule
        Berkeley Bridge & WidowX & 7 & 23.46\% \\
        FMB & Franka & 7 & 13.94\% \\
        Jaco Play   & Jaco 2 & 4 & 15.66\% \\
        Fractal & Google Robot & 7 & 46.94\% \\
        \bottomrule
    \end{tabular}
\end{table}

\begin{table}[tb]
    \centering
    \caption{Relative action prediction error on the mixed real-world test set for SmolVLA training from scratch. Co-VLA achieves the smallest gap to centralized training on average.}
    \label{tab:smolvla_scratch_mix}
    \setlength{\tabcolsep}{5.5pt}
    \begin{tabular}{@{}lccccc@{}}
    \toprule
        \textbf{Method} & \textbf{Average} & \textbf{Berkeley Bridge} & \textbf{FMB} & \textbf{Jaco Play} & \textbf{Fractal} \\
        \midrule
        Co-VLA   & \textbf{+1.3\%} & {-1.2\%} & \textbf{-1.2\%} & +1.9\% & +5.5\% \\
        FedAvg & +4.3\% & +0.0\% & +4.9\% & +7.2\% & +4.9\% \\
        DiLoCo & +2.0\% & \textbf{-2.8\%} & +4.7\% & \textbf{+1.7\%} & \textbf{+4.2\%} \\
        \bottomrule
    \end{tabular}
\end{table}

We evaluate open-loop performance using the $\ell_1$ error between predicted and ground-truth actions, as it correlates better with downstream performance than the flow-matching training loss~\cite{zheng2026xvla}. Table~\ref{tab:smolvla_scratch_mix} reports the relative error of each method regarding centralized training. Under the same computation budget, Co-VLA achieves the smallest average gap to the centralized baseline, while DiLoCo also improves upon FedAvg. On FMB, Co-VLA substantially outperforms the other federated methods. Since FMB contributes a smaller share of the training data than the other sources, one possible explanation is that the consensus formulation may help mitigate the effects of data imbalance.

\subsubsection{Closed-loop evaluation}
We evaluate closed-loop performance on an SO-101 robot arm. We collect demonstrations for four tasks: stacking, pick-and-place, sorting, and folding. The setups for the first three tasks are similar to those used in SmolVLA~\cite{shukor2025smolvla}, while folding is included as an additional task. Each task contains 50 demonstrations.
For the centralized baseline, we train SmolVLA on the combined data from all four tasks for 100,000 steps on one H100 GPU, using a batch size of 64 and a learning rate of $10^{-4}$. For federated training, we use four clients, each assigned to one task and one H100 GPU. Each client uses the same batch size and learning rate and performs 100 local steps per communication round for 1,000 rounds.
Thus, the effective total number of training samples processed across all clients per step is four times that of centralized training.
Co-VLA and DiLoCo use the same hyperparameters as in the preceding experiments.

Using LeRobot, we evaluate each method on each task over 20 trials sequentially. During evaluation, we shift object positions by approximately 1--2~cm from those in the training demonstrations while keeping them approximately consistent across methods for a fair comparison.
We repeat the complete evaluation once, yielding 40 trials per task. Table~\ref{tab:real_world_success} reports the success rates averaged over these 40 trials. The federated methods perform similarly on pick-and-place, sorting, and folding, whereas Co-VLA substantially outperforms FedAvg and DiLoCo on the stacking task. Complete recordings of the evaluation trials are available at the project website.

\begin{table}[tb]
    \centering
    \caption{Success rate in real-world experiments~(40 trials per task). Co-VLA outperforms other federated baselines.}
    \label{tab:real_world_success}
    \setlength{\tabcolsep}{6pt}
    \begin{tabular}{@{}lccccc@{}}
        \toprule
        \textbf{Method} & \textbf{Average} & \textbf{Stack} & \textbf{Pick\&Place} & \textbf{Sort} & \textbf{Fold} \\
        \midrule
        Co-VLA & \textbf{0.994} & \textbf{0.975} & \textbf{1.0} & \textbf{1.0} & \textbf{1.0} \\
        FedAvg & 0.888 & 0.625 & \textbf{1.0} & 0.925 & \textbf{1.0} \\
        DiLoCo & 0.894 & 0.675 & 0.975 & 0.925 & \textbf{1.0} \\
        \midrule
        Centralized training & 0.969 & 0.925 & 1.0 & 0.95 & 1.0 \\
        \bottomrule
    \end{tabular}
\end{table}
\section{Limitations}
Although Co-VLA keeps raw data local, this alone does not provide formal privacy guarantees, as shared model updates may still reveal information about client data. Investigating such guarantees is an interesting direction for future work. 
Furthermore, the federated training method remains slower to converge than centralized training, reflecting the challenges of decentralized and heterogeneous data. Future work could investigate more sophisticated local client optimization strategies to accelerate global convergence. Due to computational constraints, we do not perform large-scale federated pre-training. Scaling to more clients and more diverse robot embodiments with larger models remains an important direction. Our study also assumes full client participation and synchronous communication, whereas practical deployments may involve unavailable clients, stragglers, or communication failures. Extending Co-VLA to asynchronous or event-triggered ADMM is a promising direction for future work.

\section{Conclusion}
\label{sec:conclusion}
We presented Co-VLA, an application of consensus ADMM to federated VLA training under heterogeneous client data. By coupling local models to a shared global model through consensus variables and dual updates, Co-VLA provides a unified formulation for full-model training, LoRA fine-tuning, and adaptive-rank SoRA fine-tuning. 
We evaluated Co-VLA with SmolVLA and X-VLA on LIBERO and with SmolVLA on heterogeneous real-world data in open- and closed-loop settings. Across these settings, Co-VLA achieves performance comparable to centralized training. Unlike general-purpose federated baselines, Co-VLA imposes consensus directly on the low-rank factors, and it outperforms specialized federated LoRA methods on average in our experiments. In real-world experiments, Co-VLA yields a smaller average open-loop error gap to centralized training than the federated baselines and a higher success rate on the challenging stacking task. These results suggest that consensus ADMM is promising for federated VLA training when robot data are distributed and difficult to centralize.


\bibliographystyle{IEEEtran}
\bibliography{tidy_references}

\end{document}